\documentclass[sigconf, screen, review=false, timestamp=false, nonacm]{acmart}
\usepackage{multirow}
\AtBeginDocument{%
  }

\begin{document}

\title{Multi-Grained Compositional Visual Clue Learning for Image Intent Recognition}


\author{Yin Tang}
\affiliation{
  \institution{Beihang University}
  \institution{Beijing Institute for General Artificial Intelligence}
  \city{Beijing}
  \country{China}
}

\author{Jiankai Li}
\affiliation{
  \institution{Beihang University}
  \city{Beijing}
  \country{China}
}

\author{Hongyu Yang}
\affiliation{
  \institution{Beihang University}
  \city{Beijing}
  \country{China}
}

\author{Xuan Dong}
\affiliation{
  \institution{Beijing University of Post and Telecommunication}
  \city{Beijing}
  \country{China}
}

\author{Lifeng Fan}
\affiliation{
  \institution{Beijing Institute for General Artificial Intelligence}
  \city{Beijing}
  \country{China}
}

\author{Weixin Li}
\authornotemark[1]
\email{weixinli@buaa.edu.cn}
\affiliation{
  \institution{Beihang University}
  \city{Beijing}
  \country{China}
}


\begin{abstract}
In an era where social media platforms abound, individuals frequently share images that offer insights into their intents and interests, impacting individual life quality and societal stability. Traditional computer vision tasks, such as object detection and semantic segmentation, focus on concrete visual representations, while intent recognition relies more on implicit visual clues. This poses challenges due to the wide variation and subjectivity of such clues, compounded by the problem of intra-class variety in conveying abstract concepts, \textit{e.g.}  ``enjoy life''. Existing methods seek to solve the problem by manually designing representative features or building prototypes for each class from global features. However, these methods still struggle to deal with the large visual diversity of each intent category. In this paper, we introduce a novel approach named Multi-grained Compositional visual Clue Learning (MCCL) to address these challenges for image intent recognition. Our method leverages the systematic compositionality of human cognition by breaking down intent recognition into visual clue composition and integrating multi-grained features. We adopt class-specific prototypes to alleviate data imbalance. We treat intent recognition as a multi-label classification problem, using a graph convolutional network to infuse prior knowledge through label embedding correlations. Demonstrated by a state-of-the-art performance on the Intentonomy and MDID datasets, our approach advances the accuracy of existing methods while also possessing good interpretability. Our work provides an attempt for future explorations in understanding complex and miscellaneous forms of human expression.
\end{abstract}




\maketitle

\section{Introduction}
\begin{figure}[ht]
  \centering
\includegraphics[width=0.9\linewidth]{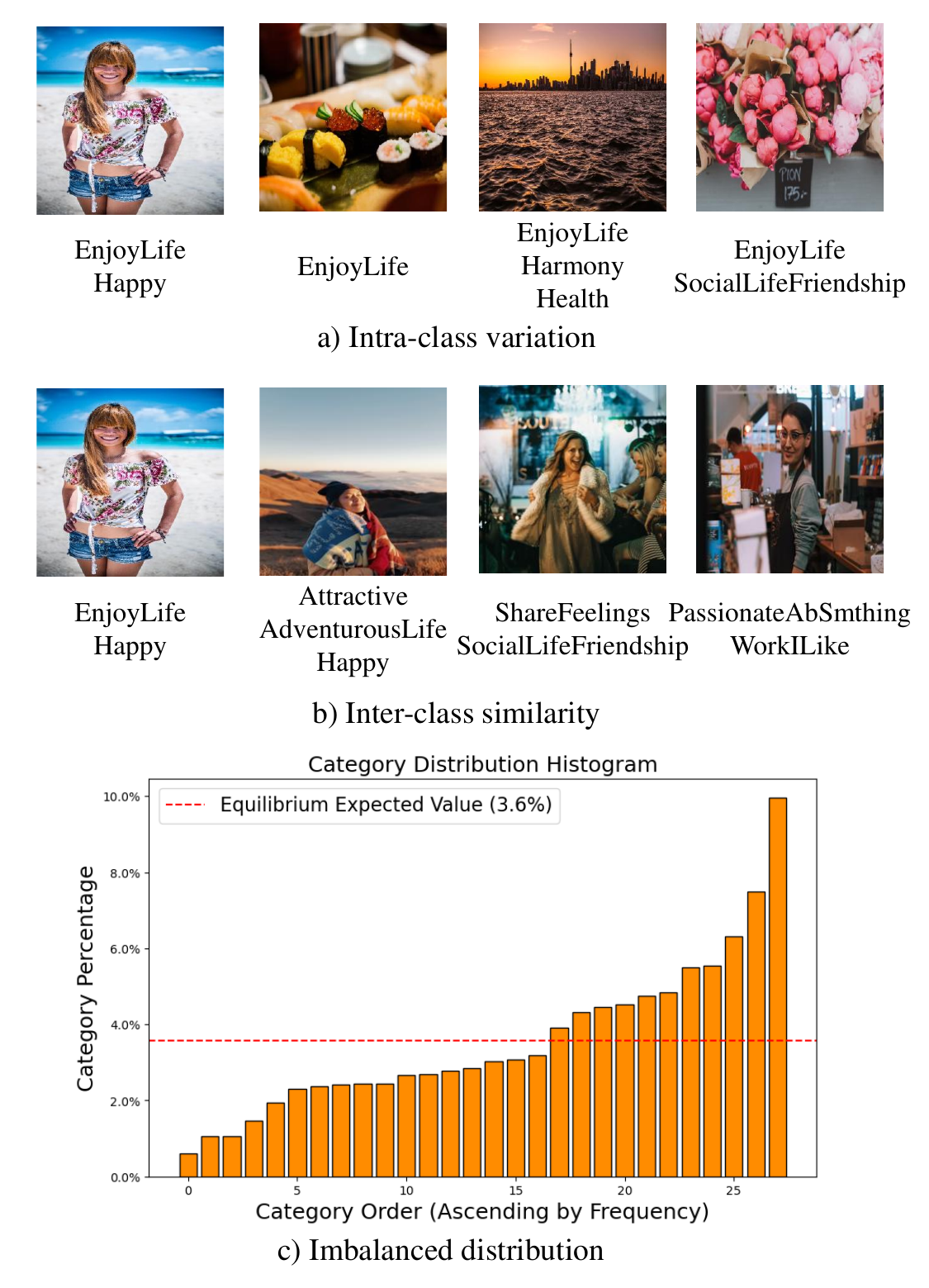}
  \caption{Main challenges in image intent recognition. a) Intra-class variation. All four images share the same intent label ``\textit{EnjoyLife}'', but their contents vary greatly. 
  b) Inter-class similarity. Though all showing a girl smiling, these images belong to distinct intent categories.
  c) Imbalanced distribution. With an equilibrium expected value of 3.6\%, more than half of the intent categories own samples less than expected, while the top one category ``\textit{NatBeauty}'' occupies nearly 10\% of the data.}
  \label{fig:example}
\end{figure}  

In today’s interconnected world, the widespread accessibility of social media makes it possible for ordinary individuals to share information and express themselves to certain audiences. 
As one of the most frequent ways that individuals choose for self-expression, images can be bonded with highly complex and implicit clues. Analyzing social media users’ intents behind the shared images can help mine their interests, quality of life, mental health status, \textit{etc.}, and thus is conducive to simultaneously monitoring and analyzing potential crisis events to ensure timely response and social stability. 
Therefore, this high-level visual comprehension task has recently garnered significant attention \cite{Jia2021, wen2023, mittal2024}.

Similar human intents can be implied by visual clues that vary tremendously. As shown by examples in Figure~\ref{fig:example} a), the same intent category ``\textit{EnjoyLife}'' can be implied from various photos, such as a girl smiling at the seaside, an exquisite dinner, a beautiful sunset, or a bunch of roses.
This is different from conventional computer vision tasks, \textit{e.g.} object detection \cite{Ren2017} or semantic segmentation \cite{Long2015}, since intent understanding relies on more abstract representations rather than those simple, explicit and concrete features of images. Meanwhile, the images sharing similar visual clues can belong to distinct intent categories. As shown in Figure~\ref{fig:example} b), all four photos have a smiling girl as the main subject, but with various contexts and thus belong to different categories. The intra-class variety, as well as the inter-class similarity in the visual clues, make it extraordinarily complex to infer high-level intent semantics with traditional classification paradigms.

In the literature, some methods \cite{Joo2014, Joo2015, Huang2016} seek to solve the problem by concentrating on certain characteristics embodied in images, \textit{e.g.,} facial appearances, gestures, scene contexts, \textit{etc.,} for inferring politician persuasive intents, which is consistent with human cognition. But when the problem extends to everyday life, it could be too wide to determine which attributes are keys to understanding the image. To this end, Wang \textit{et al.} introduce prototype learning to intent recognition, carefully selecting prototypes to represent each intent class \cite{Wang2023}. Even though their method alleviates the intra-class variety problem to some extent, its performance still cannot be guaranteed for \textit{hard} intent categories with large visual diversity. Thus, finding a way to consistently map diverse visual clues to abstract concepts in line with human cognition remains a key challenge for image intent recognition.

Considering that the powerful thinking ability of humans stems from systematic compositionality \cite{wen2023, marcus2003algebraic, lake2017building}, we observe that the complexity of the intent recognition task can be reduced by introducing \textbf{compositions of visual clues}. For instance, in Figure~\ref{fig:example} b) the combination of a smiling face and sea is connected to intent ``\textit{EnjoyLife}''. When the combination switches to a smiling face and a waitress suit, it can be inferred as ``\textit{WorkILike}''. \textbf{Multi-grained features} also contribute to intent recognition. For example, besides the higher-level features (e.g., mountains), the lower-level feature ``bright color'' also helps the recognition of intent ``\textit{CuriousAdventurousExcitingLife}''. Moreover, we notice that the intent recognition task suffers from an \textbf{imbalanced distribution}. As shown in Figure~\ref{fig:example} c), the top three categories account for more than 20\% of the dataset, while the last five small categories account for less than 5\%. Strengthening the ability to represent small categories can alleviate the bias in data for better recognition accuracy. 

\begin{figure*}[ht]
  \centering
  \includegraphics[width=\linewidth]{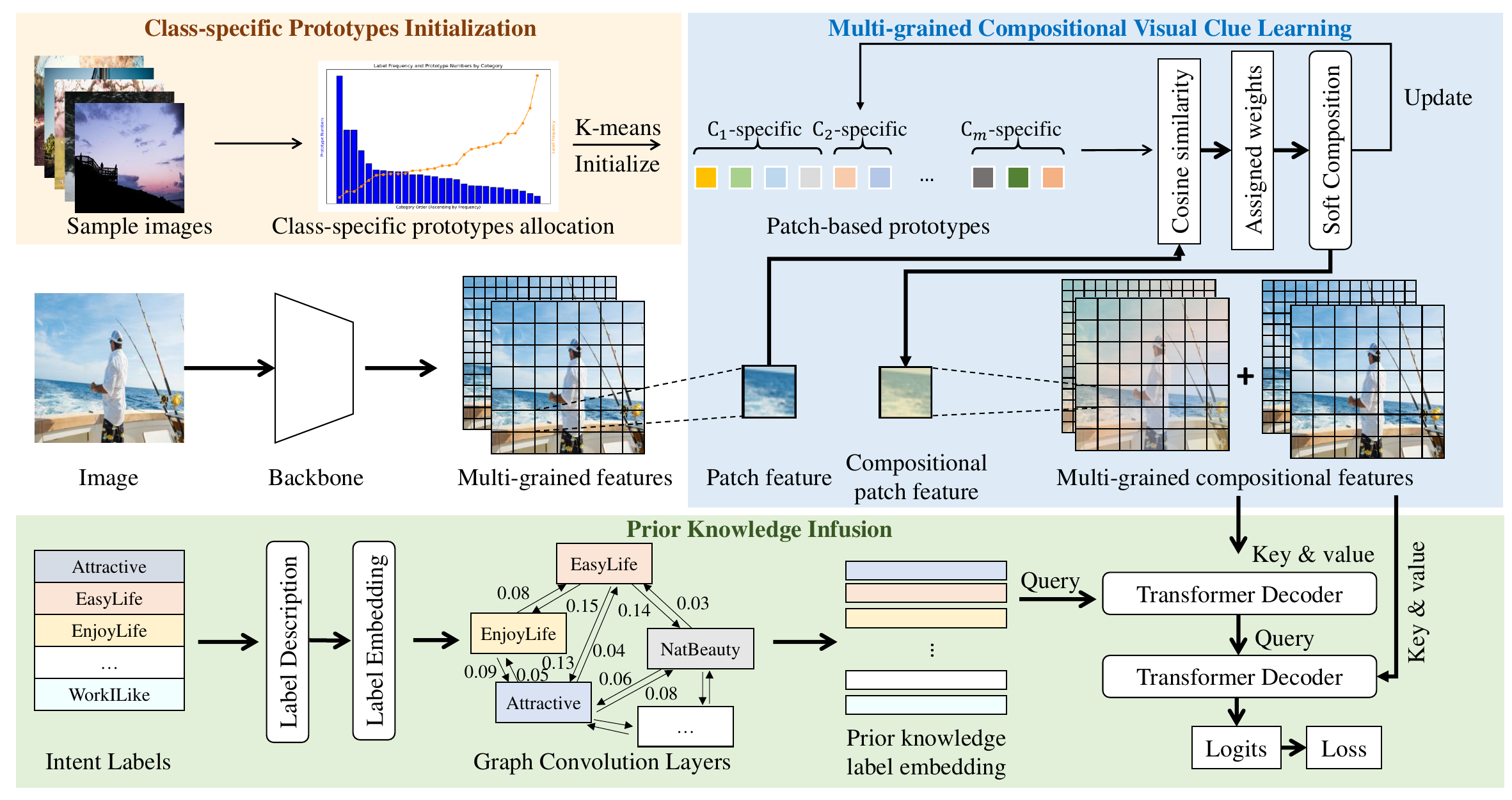}
  \caption{The framework of our proposed method. It's mainly composed of three modules, namely the Class-specific Prototype Initialization (CPI) module, the Multi-grained Compositional visual Clue learning (MCC) module, and the Prior Knowledge Infusion (PKI) module. The CPI module initializes class-specific prototypes with numbers negatively related to the proportion of categories. The MCC module learns visual representations for intent-related clues based on multi-grained features via prototype construction. The last PKI module embeds label priors through a graph convolutional network and infuses the learned knowledge into intent recognition.}
  \label{fig:framework}
\end{figure*}

Based on these insights, we propose a novel Multi-grained Compositional visual Clue Learning (MCCL) method for image intent recognition, which bridges the gap between diverse visual clues and abstract concepts. As shown in Figure~\ref{fig:framework}, first, the class-specific prototype initialization module builds prototypes for potential visual clues while enhancing the representation of small categories with a frequency-aware allocation strategy. The multi-grained compositional visual clue learning module then learns the representations by aggregating the relevant prototypes for a composition of potential clues. Finally, the prior knowledge infusion module models the correlation and semantics embedded in intent labels and classifies images into different intent categories. Specifically, each intent category is assigned several prototypes, the number of which is negatively correlated with the proportion of the category.
We obtain prototypes for potential visual clues using extracted visual features of different granularities from images in the whole training set by patch-based clustering. We learn the compositional representations for each image through the patch-prototype look-up for the compositional visual clues. A Graph Convolutional Network (GCN) is employed to infuse the intent prior information into our model by calculating correlations of label embeddings. Finally, Transformer decoder layers are utilized to classify the learned image representations and label embeddings into intent categories.

We conduct experiments on the Intentonomy \cite{Jia2021} and MDID \cite{Kruk2019} datasets and outperform state-of-the-art methods, reaching a Macro F1 score of 35.37\% and an mAP score of 37.59\% on Intentonomy and an accuracy (ACC) of 52.0\% on MDID. Ablation and analysis experiments further validate the effectiveness of our method.

Our main contributions can be summarized as follows:
\begin{itemize}
    \item We propose to learn compositions of multi-grained visual clues to solve the complex intent recognition problem.
    \item We adopt class-specific prototype initialization to deal with the imbalanced intent class distribution.
    \item We propose to jointly model visual clue learning and prior label knowledge, where these modules work together to bridge various visual clues and abstract intent concepts.
\end{itemize}

\section{Related Work}
\subsection{Image Intent Recognition}
Intent recognition seeks to decode the underlying purposes of humans. In recent years, interpreting intents through visual contents has gradually attracted attention. An array of studies has delved into discerning persuasive intents within political spheres. The Visual Persuasion dataset \cite{Joo2014} encompasses nine types of persuasive intents. Visual clues are categorized into three groups, namely facial expressions, posture and gestures, and scene backgrounds, constructing a syntactic interpretation of images. Based on this, researchers further explore the contribution of facial characteristics \cite{Joo2015}, as well as features of backgrounds and image settings \cite{Huang2016}, achieving promising results.

In research on social media user behavior, Jia \textit{et al.} construct the Intentonomy dataset \cite{Jia2021}, which contains 14,000 social media images annotated with 28 intent labels. They also investigate the significance of foreground and background information in images for different intent categories. The PIP-Net \cite{Wang2023} utilizes prototypes for clustering and matching image features, addressing the problems of intra-class variety and inter-class confusion in intent recognition. Noticing the hierarchy of intent categories, CPAD \cite{Ye2023} and HLEG \cite{Shi2023} align the relationships between image features and intent semantics with a hierarchical structure. LabCR further alleviates label shifting and blemish problems by intention calibration in augmented instance pairs, achieving promising results on the Intentonomy dataset \cite{Shi2024}.

Recognizing the similarities in visual clues can be instrumental in deducing intent from images. In this paper, we seek to represent images based on compositional clues for intent recognition. Different from PIP-Net \cite{Wang2023}, which utilizes prototypes to find the common features for each intent category, our focus shifts toward the images themselves, harnessing their inherent cognitive stimuli and multi-grained visual clues for intent prediction.

\subsection{Prototype Learning}

Prototypes can represent common patterns or features within visual data. In the field of computer vision, prototype learning is usually used in classification and clustering \cite{Lee2007, Chen2019proto}. 
By identifying prototypes, a model can reduce the dimension of the data, enabling it to focus on the most salient features and thus simplify the decision process. It can improve the interpretability of models by aligning prototypes with human-understandable concepts \cite{Chen2019proto}. 
Beyond traditional classification, prototype learning has been adapted to address specialized challenges. For instance, prototypical networks by Snell et al. \cite{snell2017prototypical} advance few-shot learning, while Zhu et al. \cite{zhu2021prototype} integrate prototypes into incremental learning frameworks. Further innovations include self-promoted prototypes for few-shot class-incremental scenarios [32] and convolutional prototype learning by Yang et al. \cite{yang2018robust}, which tackles the robustness limitations of conventional softmax layers.

In this paper, we utilize prototypes as the representation of potential visual clues for understanding intents. Class-specific prototypes are built from multi-grained features obtained from different stages of the deep network, which helps to better learn visual clues for image intent recognition.

\subsection{Multi-label Image Classification}

For the multi-label image classification problem, thanks to the publicly available large and manually-annotated datasets \textit{e.g.} ImageNet, MS-COCO \cite{Lin2014}, and PASCAL VOC \cite{Everingham2009}, many studies have been dedicated to it in the literature. 

To regulate the prediction space, researchers have attempted to capture label correlations. Deep convolutional neural networks using an approximate top-k ranking loss naturally fit the multi-label tagging problem \cite{Gong2013}. Labels transformed into vectors can be forwarded into recurrent neural networks to model relationships among labels \cite{Wang2016}. 
Noticing that attention mechanisms, which allow neural networks to selectively focus on inputs, can be employed to learn the relationships, Zhu \textit{et al.} propose a spatial regularization network and incorporate weighted attention maps to learn the semantic and spatial relationships between different labels \cite{Zhu2017}.

Considering the advantages of graph structures in expressing topological relationships, researchers also explore building graphs to represent label interrelationships \cite{Li2014, Li2016, Lee2018}. Chen \textit{et al.} \cite{Chen2019gcn} harnesses Graph Convolutional Networks (GCNs) \cite{Kipf2017} to capture and explore label dependencies and generate classifiers. 
Zhao \textit{et al.} \cite{Zhao2021} propose a Transformer-based framework, constructing complementary relationships by exploring structural relation graphs and semantic relation graphs. Liu \textit{et al.} \cite{Liu2021Q2L} leverage Transformer decoders to query the existence of labels, extracting discriminative features adaptively for different labels.

Existing studies have confirmed the effectiveness of structuring labels for multi-label image classification due to the graph structure’s ability to maintain the topology of irregular data. In this paper, we leverage the prior information with a graph convolutional network, infusing label semantic and prior probability knowledge into the learning of multi-grained images features for intent recognition.

\section{Approach}


\subsection{Overview}

Our method aims to bridge the gap between visual clues and abstract intent concepts. The framework is shown in Figure~\ref{fig:framework}. To convert multi-grained features into potential visual clues, we design prototypes to establish relationships between visual features and semantics while addressing the shortcomings of manually selecting visual clues as is done by existing methods. The prototypes represent the potential visual clues in discrete form, encoding features at each stage and allowing the model to select various regional and granular features. To enhance the representation of small categories, the class-specific initialization module assigns each category with several prototypes, the number of which is negatively correlated with the proportion of the category. Meanwhile, to leverage the prior knowledge embedded in labels, we build a graph convolutional network to embed label textual clues and regularize label relations. Finally, our model mutually recognizes intents from images by aggregating the compositional visual clues and label information with Transformer decoder layers.

\subsection{Problem Definition}
Formally, given an image $I$, the target of image intent recognition is to predict a set of intent labels $L=\{L_1, L_2, \ldots , L_C\}$, in the form of $\boldsymbol{y} = [y_1, y_2, \ldots, y_C]$, $y_i\in\{0, 1\}$, where $C$ denotes the number of intent categories, and $y_i = 1$ represents that the intent label $L_i$ can be implied from image $I$.
The model output is formed in the probability of each label, \textit{i.e.} $\boldsymbol{p} = [p_1, p_2, \ldots, p_C]$, where $p_c$ represents the probability of $y_c = 1$.

\subsection{Class-specific Prototype Initialization}
To convert the intent recognition task into compositional visual clues, prototypes are first established to offer a rule and decide which clues are embodied for intent recognition.
Image features from the training set are extracted from different stages of the backbone network (where the ResNet101 \cite{He2016} is adopted in this paper).
Then, patches are considered as the smallest unit, and clustered with K-means. Our prototype initialization employs a class-specific strategy to handle class imbalance, combining frequency-aware cluster allocation with per-class feature clustering.

Let ${\boldsymbol{F}} =[\boldsymbol{E}^1, \boldsymbol{E}^2,\ldots, \boldsymbol{E}^{P}]\in \mathbb{R}^{P\times D}$ represent the feature extracted from the pretrained backbone network, in which $P$ represents the number of patches, $P = H \times W$, and $D$ represents the dimension of feature embedding. For each class $c$, the features of samples belonging to class $c$ are picked and flattened into $\boldsymbol{E}_c \in \mathbb{R}^{(N_c \times P_s)\times D_s}$. 
Let $\mathcal{C}={1,\cdots,m}$ denote the class indices and $K$ be the total prototype budget. The initialization proceeds in two phases: cluster allocation via inverse frequency weighting and hierarchical feature clustering.
We distribute $K$ prototypes across classes using their label frequencies ${p_c}, {c\in\mathcal{C}}$. Frequencies are normalized to $\hat{p}_c = \frac{p_c}{\sum{c'} p_{c'}}$. Then, the frequencies are inversed as the initial weights $w_c = \frac{1}{\hat{p}_c}$, and normalized $\hat{w}_c = \frac{w_c}{\sum{c'} w_{c'}}$. The initial allocations are calculated as:
\begin{equation}
K_c= [\hat{w}_c \cdot K].
\end{equation}
To ensure that the number of clusters are integers, the allocations are converted through the largest remainder method.
Finally, mini-batch K-means is adopted to cluster $N_c \times H \times W$ patch features into $K_c$ initial prototypes $\hat{\boldsymbol{E}}_s^1, \hat{\boldsymbol{E}}_s^2,\ldots, \hat{\boldsymbol{E}}_s^{P}$. The prototype initialization can be conducted at any stage needed.

\begin{figure}[t]
\centering
    \includegraphics[width=\linewidth]{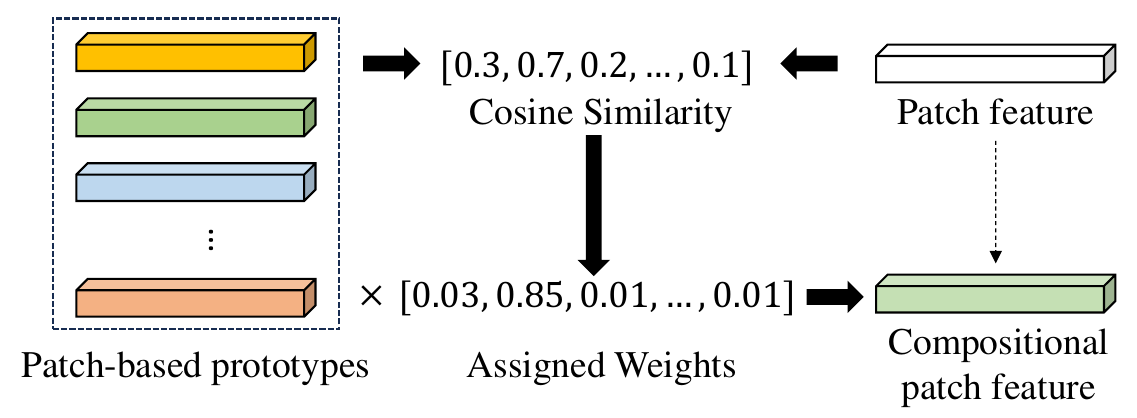}
  \caption{Illustration on patch-based feature reconstruction in the multi-grained compositional visual clue learning module. Each patch is represented as an aggregation of prototypes according to the similarity between the original patch feature and learned prototypes.}
  \label{fig:reconstruction}
\end{figure}

\subsection{Multi-Grained Compositional Visual Clue Learning}

To convert the intent recognition task into compositional visual clues, multi-grained image features are reconstructed with prototypes.
We implement a differentiable online clustering layer that simultaneously updates prototypes and generates cluster-aware representations.

\begin{table*}[ht]
\begin{tabular}{lcccccccc}
\toprule
\multirow{2}*{Methods}            & \multicolumn{4}{c}{Testing} & \multicolumn{4}{c}{Validation} \\
\cmidrule(lr){2-5} \cmidrule(lr){6-9}
                   & Macro F1 & Micro F1 & Samples F1 & mAP & Macro F1 & Micro F1 & Samples F1 & mAP \\
\midrule
Random             & 6.94     & 7.18     & 7.10       & 9.47 & 6.94     & 7.18     & 7.10       & 9.47  \\
Visual \cite{He2016}            & 23.11    & 41.23    & 42.34      & 25.01 & 24.03    & 41.36    & 42.91      & 25.98 \\
\midrule
\multicolumn{9}{l}{Multi-label image classification methods}                                                                 \\
\midrule
HiMulConE \cite{zhang2022use}         & 28.43    & 41.50    & 42.15      & 32.18 & 28.14    & 41.30    & 42.50      & 31.94 \\
MultiGuard \cite{jia2022multiguard}        & 26.87    & 38.67    & 40.06      & 36.08 & 27.74    & 41.03    & 41.58      & 37.55 \\
SST \cite{chen2022structured}               & 27.33    & 42.21    & 42.62      & 31.57 & 28.78    & 41.03    & 41.75      & 32.12 \\
Query2Label \cite{Liu2021Q2L}        & 32.12    & 44.64    & 45.15      & 34.33 & 32.77    & 44.60    & 45.40      & 35.50 \\
\midrule
\multicolumn{9}{l}{Image intent recognition methods}                                                                 \\
\midrule
LocLoss \cite{Jia2021}                & 23.98    & 31.28    & 31.39      & -     & 25.07    & 32.94    & 33.61      & -     \\
CPAD \cite{Ye2023}              & 27.39    & 40.98    & 41.12      & 29.13 & 27.37    & 41.77    & 42.68      & 29.43 \\
PIP-Net \cite{Wang2023}            & 31.51    & 44.55    & 44.73      & 35.16 & 31.80    & 46.26    & 47.02      & 34.68 \\
HLEG \cite{Shi2023}              & 32.77    & 44.69    & 45.54      & 35.93 & 35.35    & 46.34    & 47.40      & 36.86 \\
LabCR \cite{Shi2024}             & 34.63    & \textbf{48.51}    & 48.05      & 37.13 & \textbf{37.10}    & 49.04    & 49.71      & 38.86 \\
\midrule
Ours               & \textbf{35.37}    & 48.12    & \textbf{49.11}      & \textbf{37.59} & 36.65    & \textbf{49.46}    & \textbf{51.04}      & \textbf{39.52} \\
\bottomrule
\end{tabular}
    \caption{Comparison results with state-of-the-art multi-label classification and image intent recognition methods in terms of F1 scores and mAP. The best results are shown in bold.}
    \label{tab:comparison}
\end{table*}

For stage $s$, the feature extracted can be represented as ${\boldsymbol{F}}_s =[\boldsymbol{E}_s^1, \boldsymbol{E}_s^2,\ldots, \boldsymbol{E}_s^{P_s}]\in \mathbb{R}^{P_s\times D_s}$, in which $P_s$ represents the number of patches, $P_s = H_s \times W_s$, and $D_s$ represents the dimension of feature embedding in stage $s$.

We calculate the cosine similarity of each patch feature $\boldsymbol{E}_s^i$ with all the prototypes of stage $s$ as:
\begin{equation}
    sim_s^i = \frac{\boldsymbol{E}_s^i \cdot \hat{\boldsymbol{E}}_s^i}{\|\boldsymbol{E}_s^i\| \|\hat{\boldsymbol{E}}_s^i\| + \epsilon},
\end{equation}
where $\epsilon$ is a small positive constant to prevent division by zero.
Then, the weights are assigned softly with temperature $\tau$:
\begin{equation}
    W_s^i = Softmax(sim_s^i / \tau).
\end{equation}

The cluster-enhanced feature is then generated through linear combination:
\begin{equation}
    {\tilde{\boldsymbol{E}}_s^i} = \sum_{k=1}^K W_s^i \hat{\boldsymbol{E}}_s^i.
\end{equation}
Meanwhile, the prototypes are updated via momentum:
\begin{equation}
    \hat{\boldsymbol{E}}_s^{i,(t+1)} = \lambda\hat{\boldsymbol{E}}_s^i + (1-\lambda)\tilde{\boldsymbol{E}}_s^i,
\end{equation}
where $\lambda$ is the momentum hyperparameter.
 
This process can be seen as the re-representation of the patch feature, during which the image features are composed of prototypes, as shown in Figure~\ref{fig:reconstruction}. Meanwhile, the prototypes are updated, which makes clustering an online process. For an image $I$, the stage $s$ feature is reconstructed by concatenating the patch features ${\boldsymbol{\hat E}_s}$, represented as $\tilde{\boldsymbol{F}}_s = [\tilde{\boldsymbol{E}}_s^1, \tilde{\boldsymbol{E}}_s^2,\ldots, \tilde{\boldsymbol{E}}_s^{P_s}] \in \mathbb{R}^{D_s}$. Then, it is normalized and average-pooled to ${\tilde{\boldsymbol{F}}'_s} \in \mathbb{R}^{D_s}$.

\subsection{Prior Knowledge Infusion}
Previous multi-label image classification studies utilize word embedding to obtain the semantics embedded in label texts and build a structure graph to capture label correlations. The target intent labels are short words describing the main motives from social media posts. To fully leverage the semantics of labels, we first enrich the description of intents with Large Language Models (LLMs). Then, the label descriptions are converted into embeddings. We follow the work of ML-GCN \cite{Chen2019gcn} and use two stacked graph convolutional layers to learn the correlated label embedding. Through the encoding process, the label space embedding prior information is represented in the form of $\boldsymbol{E}_l \in \mathbb{R}^{m\times D}$.

After refining the representation of image features and semantic labels, we seek to map them into intent categories.
We adopt Transformer decoder layers to extract compositional feature representation with label queries. 
The label queries initialized with label embeddings are updated by decoders as:
\begin{equation}
\begin{split}
    Q_{i+1}^{(1)}&=MultiHead(\bar{Q}_{i},\bar{Q}_i,Q_i),\\
Q_{i+1}^{(2)}&=MultiHead(\bar{Q}_{i+1}^{(1)},\bar{\tilde{\boldsymbol{F}}}'_s,\tilde{\boldsymbol{F}}'_s),\\
Q_{i+1}&=FFN(Q_{i+1}^{(2)}),
\end{split}
\end{equation}
where $Q_{i+1}^{(1)}, Q_{i+1}^{(2)}$ are intermediate variables, $\bar{Q}_{i}$ represents the position embedding version of ${Q}_{i}$, $\tilde{\boldsymbol{F}}'_s$ is the compositional feature representation.
For each category $c$, the predicted score is obtained by a projection layer, the inputs of which are the concatenation of queries output by all the stages. To ensure the details in original features are kept, we adopt a residual connection:
\begin{equation}
    Q_{c}=[\hat{Q}_{L,c,1},Q_{L,c,1}, \hat{Q}_{L,c,1}, Q_{L,c,2},\ldots,\hat{Q}_{L,c,S},Q_{L,c,S}],
\end{equation}
where $L$ denotes the number of Transformer decoder layers, $S$ denotes the number of stages, $Q_{L,c,s}$ denotes the final query output from original feature $\boldsymbol{F}_s$ by the last decoder layer at stage $s$, and $\hat{Q}_{L,c,s}$ denotes the final query output from reconstructed feature $\tilde{\boldsymbol{F}}'_s$ by the last decoder layer at stage $s$.
Finally, the task is treated as $m$ binary-classification problems. The probability of the image belongs to intent category is calculated with a projection layer with a Sigmoid layer followed:
\begin{equation}
    \hat {\boldsymbol{p}}_c = Sigmoid(WQ_{c}+b),
\end{equation}
where $W$ and $b$ denote the weight and bias of the projection layer, respectively.

\section{Experiments}
To prove the effectiveness of our approach, we conduct extensive experiments on the widely used and publicly available social media intent datasets, namely Intentonomy \cite{Jia2021} and MDID \cite{Kruk2019} datasets. In this section, we first introduce the dataset, evaluation metrics, and detailed implementation of our experiments. Then, we present the comparison experiments, ablation studies, and qualitative analysis of our designed method.

\subsection{Datasets}
\subsubsection{Intentonomy}
The Intentonomy \cite{Jia2021} dataset is built up of 14,455 high-resolution photos from the website Unsplash, which are respectively split into 12,740, 498 and 1,217 images for training, validation and testing. Each photo is annotated with one or more intent labels from a total set of 28 labels which are organized in a 4-level hierarchy. The intent labels form a set of motives from social media posts, covering a wide range of everyday life scenes.

\subsubsection{MDID}
The MDID \cite{Kruk2019} dataset contains 1,299 Instagram posts labeled for authorial intents and contextual and semiotic relationships between the image and text. It is a multi-class dataset, where each sample is labeled from 8 intent labels generalized and elaborated from existing rhetorical categories.

\subsection{Evaluation Metrics}
For the multi-label classification task, we employ {\itshape Macro F1}, {\itshape Sample F1}, and {\itshape Micro F1} scores to measure the model performance since the distribution might be imbalanced across the categories. The {\itshape Macro F1} score indicates the ability to handle class imbalance and measures performance on per-class classification. The {\itshape Sample F1} score evaluates the model's performance on each sample by considering the average F1 score per class weighted by support — the number of true instances for each label. The {\itshape Micro F1} score implies the precision-recall balance for each individual classification. We also adopt {\itshape mAP} to measure the performance, which is a common criterion in the multi-label classification task. For multi-class classification tasks, we report the classification accuracy (ACC) as well as the area under the ROC curve (AUC) using the macro-average across all classes.

\subsection{Implementation Details}
Our approach utilizes the ResNet101 model pre-trained on the ImageNet dataset \cite{deng2009imagenet} as the backbone. The last 2 stages of features are extracted and represented. The input images are resized to the resolution of 224 $\times$ 224 and cutout \cite{devries2017improved} with a factor of 0.5 and True-wight-decay \cite{loshchilov2017decoupled} of $1e^{-2}$. Moreover, we normalize input images with mean $[0, 0, 0]$ and std $[1, 1, 1]$ and use RandAugment \cite{cubuk2020randaugment} for augmentation. The number of all prototypes is 2,048, the hyperparameter $\tau$ is set to 0.1, and $\lambda$ is set to 0.99999. The labels are enriched in text with GPT-3.5 Turbo and embedded with text-embedding-ada-002. We adopt 2 Transformer decoder layers for classification. As for the training details, we use the Adam optimizer \cite{kingma2014adam} and 1-cycle policy \cite{cubuk2020randaugment} with a maximal learning rate of $1e^{-4}$. Following common practices, we apply an exponential moving average (EMA) to model parameters with a decay of 0.9997. The prediction loss is calculated by simplified asymmetric loss \cite{ridnik2021asymmetric}, with parameters $\gamma^+=0$, and $\gamma^-=2$.

\subsection{Comparison Experiments}
We conduct the comparison of our method with state-of-the-art models on the Intentonomy and MDID datasets, respectively. 

\subsubsection{Results on Intentonomy}
We compare our method with traditional multi-label classification methods and image intent recognition methods. The comparison results are shown in Table~\ref{tab:comparison}, which shows that our MCCL method can outperform the previous multi-label classification paradigm, gaining an improvement of 5.57\% of Macro F1 compared to Query2Label \cite{Liu2021Q2L}.

The results also offer a clear comparative analysis between our MCCL method and the current state-of-the-art models on the Intentonomy dataset. The compared models include LocLoss \cite{Jia2021}, CPAD \cite{Ye2023}, PIP-Net \cite{Wang2023}, HLEG \cite{Shi2023} and LabCR \cite{Shi2024}. Our MCCL method outperforms most of the primary methods listed, marking substantial improvements across the board. Most impressive is the 35.37\% Macro F1 score, which reflects the model's improved ability to identify each class with a fair performance irrespective of the class's frequency in the dataset, which is crucial for tasks with a diverse set of classes like intent classification. This boost suggests that our MCCL is more adept at handling a varying number of samples for different classes, which is particularly challenging in datasets with class imbalance. And with a 1.06\% increase in Samples F1 score and a 0.46\% increase in mAP, our MCCL method demonstrates a finer granularity in performance for instance-level classification. This comprehensive performance gain is attributed to the robust design of our MCCL method, which effectively captures the complex and nuanced features necessary for accurate intent classification. These comparisons underscore the effectiveness of the MCCL method in understanding and categorizing intentions better than its contemporaries, showcasing its potential as a sophisticated tool in the domain of intent recognition.

\subsubsection{Results on MDID}
Due to the restrictions on releasing image data collected from social media websites, only the deep features for each image (obtained from the last layer of ResNet-18) are released for MDID dataset. We encode feature representation from the given features only, which operates as an additional branch. We perform 5-fold cross-validation and give the average performance due to the small size of the dataset.

\begin{table}[t]
\centering
  \begin{tabular}{ccc}
    \toprule
    Methods&ACC& AUC\\
    \midrule
    Chance& 28.1&50.0\\
    DCNN \cite{Kruk2019} & 42.9&  73.8\\
    TRA (60\%) \cite{gonzaga2021multimodal} & 51.1&  81.9\\
    TRA (80\%) \cite{gonzaga2021multimodal} & 54.4&  84.4\\
    \midrule 
    Ours & 52.0& 83.8\\
    \bottomrule
\end{tabular}
\caption{Comparison with other methods on the MDID dataset in terms of accuracy (ACC) and area under the ROC curve (AUC). For TRA, 60\% and 80\% represent the presence percentages of textual information.}
  \label{tab:compare-MDID}
\end{table}

The comparison results are shown in Table~\ref{tab:compare-MDID}. Compared to the baseline (DCNN \cite{Kruk2019} method), which is a simple classifier appended on extracted features, our additional branch of feature representation greatly improves the performance. 
For TRA, 60\% and 80\% represent the presence percentages of textual information. Our methods outperforms the result of TRA (60\%) without any text information. Results on the MDID dataset show that even one stage of our feature representation method is effective. It also demonstrates that our proposed feature representation method has the flexibility of being adopted in various backbones.

\subsection{Ablation Studies}
\subsubsection{Ablation Study of Components}
We conduct ablation studies to explore the effectiveness of our designed class-specific prototype initialization module, the multi-grained compositional visual clue learning module, and the prior knowledge infusion module. The baseline is Query2Label \cite{Liu2021Q2L} with two decoder layers. The results are shown in Table~\ref{tab:ablation-module}.

\begin{table*}[ht]
\begin{tabular}{lcccccccc}
\toprule
\multirow{2}*{Model}          & \multicolumn{4}{c}{Testing} & \multicolumn{4}{c}{Validation} \\
\cmidrule(lr){2-5} \cmidrule(lr){6-9}
               & Macro F1 & Micro F1 & Samples F1 & mAP & Macro F1 & Micro F1 & Samples F1 & mAP \\
\midrule
Baseline       & 32.12    & 44.64    & 45.15      & 34.33 & 32.77    & 44.60    & 45.40      & 35.50 \\
+ PKI           & 34.70    & 45.18    & 43.93      & 36.15 & 35.43    & 43.28    & 44.84      & 38.69 \\
+ PKI + MCC       & 34.79    & \textbf{48.54}    & 48.73      & \textbf{38.17} & 35.66    & 48.97    & 50.00      & 39.35 \\
+ PKI + MCC + CPI   & \textbf{35.37}    & 48.12    & \textbf{49.11}      & 37.59 & \textbf{36.65}    & \textbf{49.46}    & \textbf{51.04}      & \textbf{39.52} \\
\bottomrule
\end{tabular}
  \caption{Ablation study of Class-specific Prototype Initialization (CPI) module, the Multi-grained Compositional visual Clue learning (MCC) module, and the Prior Knowledge Infusion (PKI) module on the Intentonomy dataset.}
    \label{tab:ablation-module}
\end{table*}

Compared to the baseline, prior knowledge infusion further improves the performance of multi-label classification by offering prior label knowledge. The increase of Macro F1 and mAP suggests KPI integration enhances class-wise discrimination and ranking stability. 
The multi-grained compositional visual clue learning module gains significant gains on Micro F1 and Samples F1.
Testing mAP jumps from 36.15\% to 38.17\%, showing improved ability on the image intent recognition task.
Macro F1 and Validation mAP peak with class-specific prototype initialization module, indicating better handling of class imbalance.
Testing mAP drops slightly, resulting from minor trade-offs between class-specific constraints and classification accuracy.

Moreover, we conduct experiments to show the effectiveness of multi-grained features, as shown in Table~\ref{tab:ablation-layers}. Leveraging the features from the last two layers outperforms the last layer only, which could result from lower-level features such as brightness. However, the performance decreases when coarse features from lower layers are added. Considering that the first and second layers contain redundant low-level patterns, the compositions are more challenging to learn and thus cause the decrease.

\begin{table}[t]
\centering
  \begin{tabular}{ccccccccc}
    \toprule
     \multicolumn{4}{c}{Feature Layers}&\multirow{2}*{Macro F1}&\multirow{2}*{Samples F1}&\multirow{2}*{Micro F1}&\multirow{2}*{mAP}\\
     \cmidrule(lr){1-4}
     L1&L2&L3&L4&&&&\\
    \midrule
     &&&\checkmark&   36.44  & 47.94  & 49.69 & 39.37\\
     &&\checkmark&\checkmark&  \textbf{36.65}   &  \textbf{49.46} & \textbf{51.04} & 39.52\\
     &\checkmark&\checkmark&\checkmark&  36.49   & 47.43  & 49.10 & \textbf{39.73}\\
     \checkmark&\checkmark&\checkmark&\checkmark&  34.83   & 45.15  & 46.16 & 38.74\\
    \bottomrule
\end{tabular}
  \caption{Ablation study of feature layers for compositional visual task learning on the validation set of the Intentonomy dataset. L1, L2, L3, and L4 represent the first, second, third, and fourth layer of output from the backbone, respectively.}
    \label{tab:ablation-layers}
    \vspace{-0.5cm}
\end{table}

\subsubsection{Ablation study of Hyperparameters}
We vary the values of the number of prototypes $K$, and the results are shown in Table~\ref{tab:ablation-size}. The ablation study is conducted on the validation set of the Intentonomy dataset. The size of prototypes determines the capacity of containing various visual clues. The overall performance gets better when the size increases. However, the expansion of size is not necessarily positively related to better performance, as it might lead to the generation of useless prototypes that actually have few features retrieved.

\begin{table}[t]
    \centering
    \begin{tabular}{ccccc}
    \toprule
         $K$&  Macro F1&  Micro F1& Samples F1& mAP\\
         \midrule
         128&  33.79&  48.53& 49.50& 37.65\\
         256&  36.98&  48.91& 50.39& 39.20\\
         512&  35.39&  48.64& 50.05& 38.92\\
         1024&  34.15&  48.71& 49.57& 39.44\\
         2048&  36.65&  49.46& 51.04& 39.52\\
         4096&  36.24&  48.49& 50.70& 39.22\\
         \bottomrule
    \end{tabular}
    \caption{Ablation study of prototype numbers on the validation set of Intentonomy dataset.}
    \label{tab:ablation-size}
    \vspace{-0.5cm}
\end{table}

The value of momentum $\lambda$ decides the extent to which the original prototypes are retained when updating prototypes. The results of the ablation study of momentum on the validation set of Intentonomy dataset are shown in Table~\ref{tab:ablation-momentum}. When $\lambda$ is relatively small, the prototypes vary quickly from the initial values, thus gaining poor performance. When $\lambda = 1.0$, the prototypes don't update during the training progress, which is restricted to the features learned by the pre-trained backbone, demonstrating the effectiveness of online clustering.

\begin{table}[t]
    \centering
    \begin{tabular}{ccccc}
    \hline
         $\lambda$&  Macro F1&  Micro F1& Samples F1 & mAP\\
         \hline 
         0.9&  35.14&  47.40& 49.92& 39.50\\
         0.99&  35.38&  49.76& 51.30& 38.25\\
         0.999&  36.37&  47.30& 49.86& 38.86\\
         0.9999&  34.90&  47.88& 49.83& 38.02\\
         0.99999&  36.65&  49.46& 51.04& 39.52\\
         1.0& 35.01 & 47.70 & 49.88 & 38.38\\
         \hline
    \end{tabular}
    \caption{Ablation study of momentum value $\lambda$ when updating prototypes on the validation set of Intentonomy dataset.}
    \label{tab:ablation-momentum}
    \vspace{-0.5cm}
\end{table}

\subsection{Qualitative Analysis}
\subsubsection{Effectiveness of Visual Clue Learning}
\begin{figure}[t]
\centering
    \includegraphics[width=\linewidth]{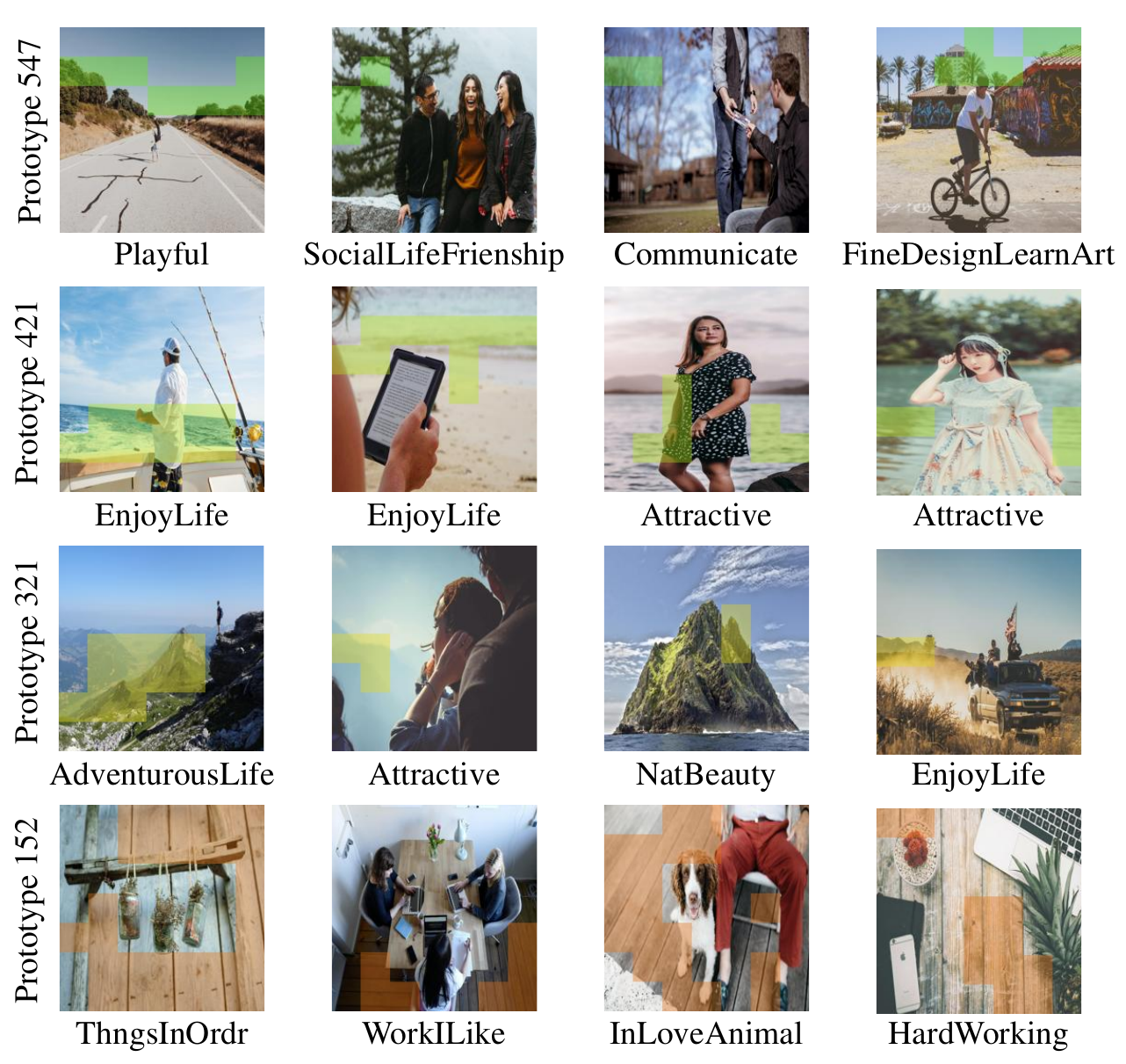}
  \caption{Examples of image patches in various intent categories highlight with different prototypes.}
  \label{fig:prototype-examples}
\end{figure}

Visual clue learning is designed to represent features from the patch perspectives with prototypes, and each prototype embeds a certain visual clue that is common and significant to intent recognition. Figure~\ref{fig:prototype-examples} presents examples with patches highlighted according to codes of inducted clues. It can be seen that prototype 547 occurs when the patches are associated with ``tree'', while prototype 421 is connected with ``water''. It indicates that our model successfully inducts clues with semantic meanings, which contributes to visual clue composition for intent recognition.

\begin{figure}[t]
    \centering
    \includegraphics[width=0.9\linewidth]{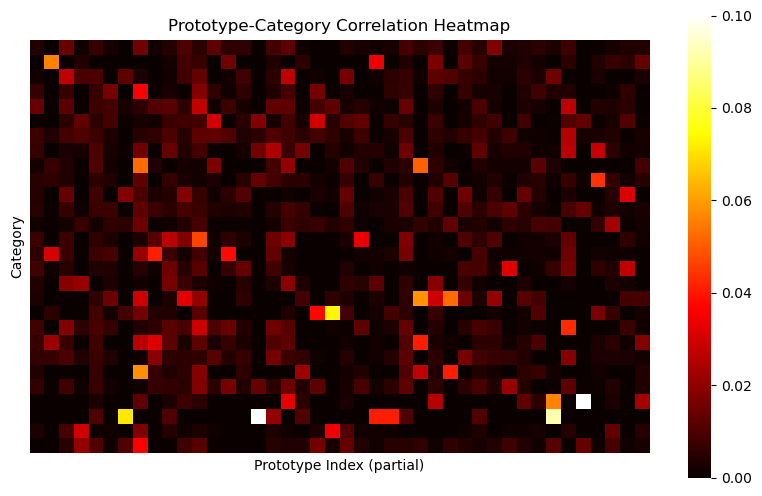}
    \caption{Heat map of prototype-category correlation. The brightness of squares is positively correlated to the frequency that the features of an intent category are retrieved to a certain prototype.}
    \label{fig:prototype-category}
\end{figure}

Besides, we visualize the heat map of prototype-category correlation in Figure~\ref{fig:prototype-category}. A brighter square represents stronger connections between the learned prototype and intent category. The result is consistent with what we expect, that there are isolated squares horizontally and vertically, which means that an intent category is related to certain prototypes, while each prototype contributes to the intent category distinctly.

\begin{figure}[t]
  \centering
  \includegraphics[width=\linewidth]{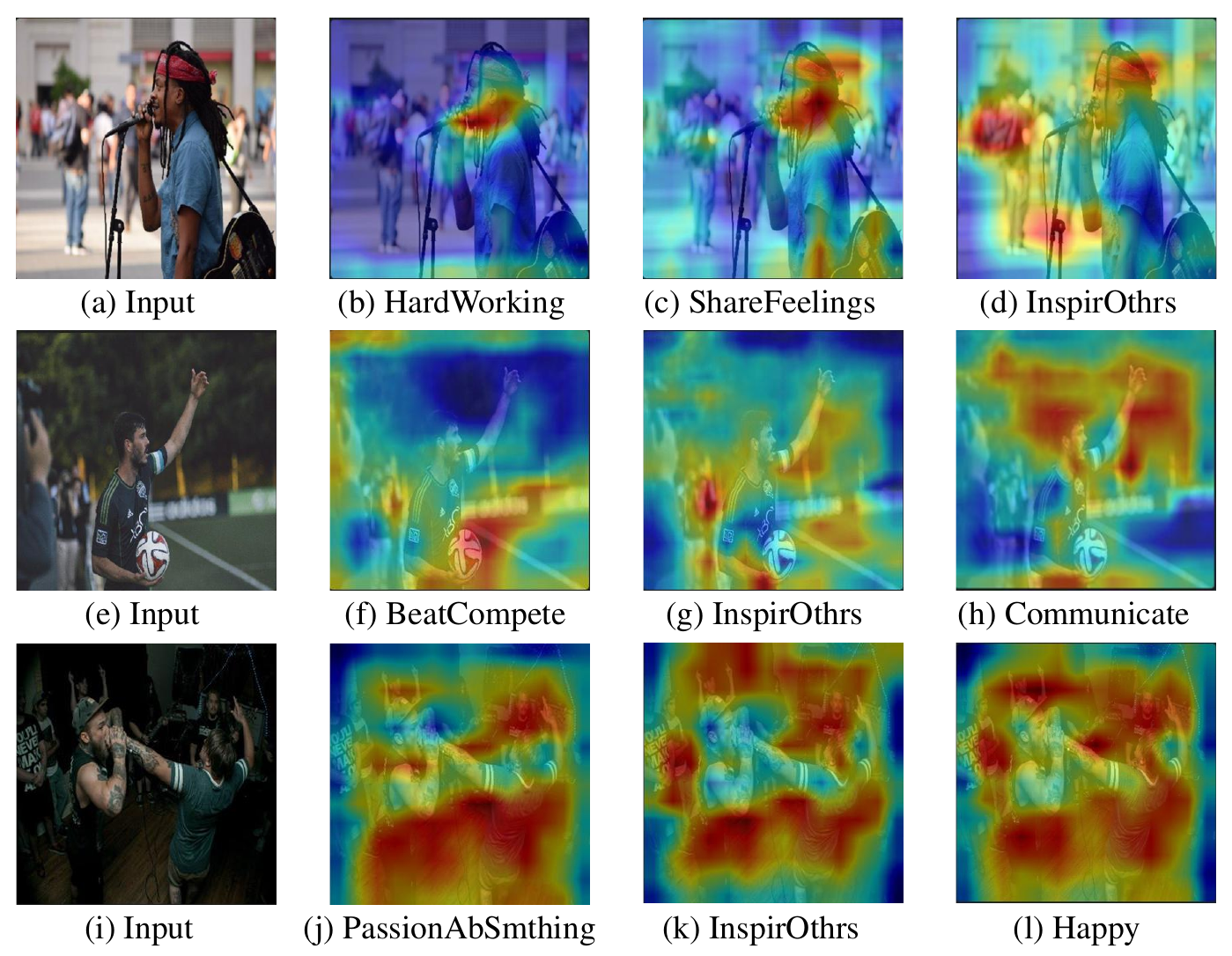}
  \caption{Activation map visualization of different intent categories for the same input image. Distinct visual elements are concentrated for different intent categories, which demonstrates that various visual clues are composed for recognition as expected.}
  \label{fig:cam}
\end{figure}

\subsubsection{Visualization of Attention Maps}
To validate the interpretability and multi-label classification capacity of our Multi-grained Compositional visual Clue Learning (MCCL) model, we employ Class Activation Mapping (CAM) \cite{selvaraju2017} to highlight regions of an image most influential in predicting a specific intent category. As illustrated in Figure~\ref{fig:cam}, these heatmaps reveal how MCCL concentrates on distinct visual elements for different intent categories.
For instance, in the first row of Figure~\ref{fig:cam}, which depicts a singer performing in a crowded public space, the model exhibits intent-specific spatial reasoning. For the intent ``hard-working'', the model focuses on the microphone held by the singer; for ``share feelings'', the singer himself is concentrated; while for ``inspire others'', the passengers, as well as the singer, are included for classification. It demonstrates that our model focuses on distinct regions in images to recognize different intents.

\section{Conclusions}
In this paper, we propose a Multi-grained Compositional visual Clue Learning (MCCL) model for image intent recognition.
Our framework formalizes intent recognition as a composition process, where multi-grained features are represented by aggregations of prototypes for a composition of visual clues.
Acknowledging the imbalanced data distribution in intent datasets, we design class-specific prototype initialization with a frequency-aware allocation strategy.
To leverage the correlation and semantics embedded in intent labels, we use a graph convolutional network to infuse the prior knowledge.
Experiments on two intent datasets (Intentonomy and MDID) prove the effectiveness of the proposed method. We gain improvements in performance compared with state-of-the-art methods.
Qualitative case studies reveal that MCCL enhances the interpretability of image intent recognition.

\bibliographystyle{ACM-Reference-Format}
\bibliography{sample-base}


\begin{thebibliography}{43}


\ifx \showCODEN    \undefined \def \showCODEN     #1{\unskip}     \fi
\ifx \showISBNx    \undefined \def \showISBNx     #1{\unskip}     \fi
\ifx \showISBNxiii \undefined \def \showISBNxiii  #1{\unskip}     \fi
\ifx \showISSN     \undefined \def \showISSN      #1{\unskip}     \fi
\ifx \showLCCN     \undefined \def \showLCCN      #1{\unskip}     \fi
\ifx \shownote     \undefined \def \shownote      #1{#1}          \fi
\ifx \showarticletitle \undefined \def \showarticletitle #1{#1}   \fi
\ifx \showURL      \undefined \def \showURL       {\relax}        \fi
\providecommand\bibfield[2]{#2}
\providecommand\bibinfo[2]{#2}
\providecommand\natexlab[1]{#1}
\providecommand\showeprint[2][]{arXiv:#2}

\bibitem[Chen et~al\mbox{.}(2019a)]%
        {Chen2019proto}
\bibfield{author}{\bibinfo{person}{Chaofan Chen}, \bibinfo{person}{Oscar Li}, \bibinfo{person}{Chaofan Tao}, \bibinfo{person}{Alina~Jade Barnett}, \bibinfo{person}{Jonathan Su}, {and} \bibinfo{person}{Cynthia Rudin}.} \bibinfo{year}{2019}\natexlab{a}.
\newblock \showarticletitle{This Looks Like That: Deep Learning for Interpretable Image Recognition}. In \bibinfo{booktitle}{\emph{Neural Information Processing Systems}}. Article \bibinfo{articleno}{801}, \bibinfo{numpages}{12}~pages.
\newblock


\bibitem[Chen et~al\mbox{.}(2022)]%
        {chen2022structured}
\bibfield{author}{\bibinfo{person}{Tianshui Chen}, \bibinfo{person}{Tao Pu}, \bibinfo{person}{Hefeng Wu}, \bibinfo{person}{Yuan Xie}, {and} \bibinfo{person}{Liang Lin}.} \bibinfo{year}{2022}\natexlab{}.
\newblock \showarticletitle{Structured Semantic Transfer for Multi-label Recognition with Partial Labels}. In \bibinfo{booktitle}{\emph{AAAI Conference on Artificial Intelligence}}, Vol.~\bibinfo{volume}{36}. \bibinfo{pages}{339--346}.
\newblock


\bibitem[Chen et~al\mbox{.}(2019b)]%
        {Chen2019gcn}
\bibfield{author}{\bibinfo{person}{Zhao-Min Chen}, \bibinfo{person}{Xiu-Shen Wei}, \bibinfo{person}{Peng Wang}, {and} \bibinfo{person}{Yanwen Guo}.} \bibinfo{year}{2019}\natexlab{b}.
\newblock \showarticletitle{Multi-Label Image Recognition With Graph Convolutional Networks}. In \bibinfo{booktitle}{\emph{IEEE Conference on Computer Vision and Pattern Recognition}}. \bibinfo{pages}{5172--5181}.
\newblock


\bibitem[Cubuk et~al\mbox{.}(2020)]%
        {cubuk2020randaugment}
\bibfield{author}{\bibinfo{person}{Ekin~D Cubuk}, \bibinfo{person}{Barret Zoph}, \bibinfo{person}{Jonathon Shlens}, {and} \bibinfo{person}{Quoc~V Le}.} \bibinfo{year}{2020}\natexlab{}.
\newblock \showarticletitle{Randaugment: Practical automated data augmentation with a reduced search space}. In \bibinfo{booktitle}{\emph{2020 IEEE/CVF Conference on Computer Vision and Pattern Recognition Workshops}}. \bibinfo{pages}{702--703}.
\newblock


\bibitem[Deng et~al\mbox{.}(2009)]%
        {deng2009imagenet}
\bibfield{author}{\bibinfo{person}{Jia Deng}, \bibinfo{person}{Wei Dong}, \bibinfo{person}{Richard Socher}, \bibinfo{person}{Li-Jia Li}, \bibinfo{person}{Kai Li}, {and} \bibinfo{person}{Li Fei-Fei}.} \bibinfo{year}{2009}\natexlab{}.
\newblock \showarticletitle{Imagenet: A Large-scale Hierarchical Image Database}. In \bibinfo{booktitle}{\emph{IEEE Conference on Computer Vision and Pattern Recognition}}. \bibinfo{pages}{248--255}.
\newblock


\bibitem[DeVries and Taylor(2017)]%
        {devries2017improved}
\bibfield{author}{\bibinfo{person}{Terrance DeVries} {and} \bibinfo{person}{Graham~W Taylor}.} \bibinfo{year}{2017}\natexlab{}.
\newblock \showarticletitle{Improved Regularization of Convolutional Neural Networks with Cutout}.
\newblock \bibinfo{journal}{\emph{arXiv preprint arXiv:1708.04552}} (\bibinfo{year}{2017}).
\newblock


\bibitem[Everingham et~al\mbox{.}(2009)]%
        {Everingham2009}
\bibfield{author}{\bibinfo{person}{Mark Everingham}, \bibinfo{person}{Luc Van~Gool}, \bibinfo{person}{Christopher K.~I. Williams}, \bibinfo{person}{John Winn}, {and} \bibinfo{person}{Andrew Zisserman}.} \bibinfo{year}{2009}\natexlab{}.
\newblock \showarticletitle{The Pascal Visual Object Classes (VOC) Challenge}.
\newblock \bibinfo{journal}{\emph{International Journal of Computer Vision}} \bibinfo{volume}{88}, \bibinfo{number}{2} (\bibinfo{date}{Sept.} \bibinfo{year}{2009}), \bibinfo{pages}{303–338}.
\newblock
\showISSN{1573-1405}


\bibitem[Gong et~al\mbox{.}(2013)]%
        {Gong2013}
\bibfield{author}{\bibinfo{person}{Yunchao Gong}, \bibinfo{person}{Yangqing Jia}, \bibinfo{person}{Thomas Leung}, \bibinfo{person}{Alexander Toshev}, {and} \bibinfo{person}{Sergey Ioffe}.} \bibinfo{year}{2013}\natexlab{}.
\newblock \showarticletitle{Deep Convolutional Ranking for Multilabel Image Annotation}.
\newblock \bibinfo{journal}{\emph{arXiv preprint arXiv:1312.4894}} (\bibinfo{year}{2013}).
\newblock


\bibitem[Gonzaga et~al\mbox{.}(2021)]%
        {gonzaga2021multimodal}
\bibfield{author}{\bibinfo{person}{Victor~Machado Gonzaga}, \bibinfo{person}{Nils Murrugarra-Llerena}, {and} \bibinfo{person}{Ricardo Marcacini}.} \bibinfo{year}{2021}\natexlab{}.
\newblock \showarticletitle{Multimodal intent classification with incomplete modalities using text embedding propagation}. In \bibinfo{booktitle}{\emph{Brazilian Symposium on Multimedia and the Web}}. \bibinfo{pages}{217--220}.
\newblock


\bibitem[He et~al\mbox{.}(2016)]%
        {He2016}
\bibfield{author}{\bibinfo{person}{Kaiming He}, \bibinfo{person}{Xiangyu Zhang}, \bibinfo{person}{Shaoqing Ren}, {and} \bibinfo{person}{Jian Sun}.} \bibinfo{year}{2016}\natexlab{}.
\newblock \showarticletitle{Deep Residual Learning for Image Recognition}. In \bibinfo{booktitle}{\emph{IEEE Conference on Computer Vision and Pattern Recognition}}. \bibinfo{pages}{770--778}.
\newblock


\bibitem[Huang and Kovashka(2016)]%
        {Huang2016}
\bibfield{author}{\bibinfo{person}{Xinyue Huang} {and} \bibinfo{person}{Adriana Kovashka}.} \bibinfo{year}{2016}\natexlab{}.
\newblock \showarticletitle{Inferring Visual Persuasion via Body Language, Setting, and Deep Features}. In \bibinfo{booktitle}{\emph{IEEE Conference on Computer Vision and Pattern Recognition Workshops}}. \bibinfo{pages}{778--784}.
\newblock


\bibitem[Jia et~al\mbox{.}(2022)]%
        {jia2022multiguard}
\bibfield{author}{\bibinfo{person}{Jinyuan Jia}, \bibinfo{person}{Wenjie Qu}, {and} \bibinfo{person}{Neil Gong}.} \bibinfo{year}{2022}\natexlab{}.
\newblock \showarticletitle{Multiguard: Provably Robust Multi-label Classification against Adversarial Examples}. In \bibinfo{booktitle}{\emph{Neural Information Processing Systems}}, Vol.~\bibinfo{volume}{35}. \bibinfo{pages}{10150--10163}.
\newblock


\bibitem[Jia et~al\mbox{.}(2021)]%
        {Jia2021}
\bibfield{author}{\bibinfo{person}{Menglin Jia}, \bibinfo{person}{Zuxuan Wu}, \bibinfo{person}{Austin Reiter}, \bibinfo{person}{Claire Cardie}, \bibinfo{person}{Serge Belongie}, {and} \bibinfo{person}{Ser-Nam Lim}.} \bibinfo{year}{2021}\natexlab{}.
\newblock \showarticletitle{Intentonomy: a Dataset and Study towards Human Intent Understanding}. In \bibinfo{booktitle}{\emph{IEEE Conference on Computer Vision and Pattern Recognition}}. \bibinfo{pages}{12981--12991}.
\newblock


\bibitem[Joo et~al\mbox{.}(2014)]%
        {Joo2014}
\bibfield{author}{\bibinfo{person}{Jungseock Joo}, \bibinfo{person}{Weixin Li}, \bibinfo{person}{Francis~F. Steen}, {and} \bibinfo{person}{Song-Chun Zhu}.} \bibinfo{year}{2014}\natexlab{}.
\newblock \showarticletitle{Visual Persuasion: Inferring Communicative Intents of Images}. In \bibinfo{booktitle}{\emph{IEEE Conference on Computer Vision and Pattern Recognition}}. \bibinfo{pages}{216--223}.
\newblock


\bibitem[Joo et~al\mbox{.}(2015)]%
        {Joo2015}
\bibfield{author}{\bibinfo{person}{Jungseock Joo}, \bibinfo{person}{Francis~F. Steen}, {and} \bibinfo{person}{Song-Chun Zhu}.} \bibinfo{year}{2015}\natexlab{}.
\newblock \showarticletitle{Automated Facial Trait Judgment and Election Outcome Prediction: Social Dimensions of Face}. In \bibinfo{booktitle}{\emph{IEEE International Conference on Computer Vision}}. \bibinfo{pages}{3712--3720}.
\newblock


\bibitem[Kingma and Ba(2014)]%
        {kingma2014adam}
\bibfield{author}{\bibinfo{person}{Diederik~P Kingma} {and} \bibinfo{person}{Jimmy Ba}.} \bibinfo{year}{2014}\natexlab{}.
\newblock \showarticletitle{Adam: A Method for Stochastic Optimization}.
\newblock \bibinfo{journal}{\emph{arXiv preprint arXiv:1412.6980}} (\bibinfo{year}{2014}).
\newblock


\bibitem[Kipf and Welling(2017)]%
        {Kipf2017}
\bibfield{author}{\bibinfo{person}{Thomas~N. Kipf} {and} \bibinfo{person}{Max Welling}.} \bibinfo{year}{2017}\natexlab{}.
\newblock \showarticletitle{Semi-Supervised Classification with Graph Convolutional Networks}. In \bibinfo{booktitle}{\emph{International Conference on Learning Representations}}. \bibinfo{pages}{1--10}.
\newblock


\bibitem[Kruk et~al\mbox{.}(2019)]%
        {Kruk2019}
\bibfield{author}{\bibinfo{person}{Julia Kruk}, \bibinfo{person}{Jonah Lubin}, \bibinfo{person}{Karan Sikka}, \bibinfo{person}{Xiao Lin}, \bibinfo{person}{Dan Jurafsky}, {and} \bibinfo{person}{Ajay Divakaran}.} \bibinfo{year}{2019}\natexlab{}.
\newblock \showarticletitle{Integrating Text and Image: Determining Multimodal Document Intent in Instagram Posts}. In \bibinfo{booktitle}{\emph{2019 Conference on Empirical Methods in Natural Language Processing and the 9th International Joint Conference on Natural Language Processing}}. \bibinfo{pages}{4622--4632}.
\newblock


\bibitem[Lake et~al\mbox{.}(2017)]%
        {lake2017building}
\bibfield{author}{\bibinfo{person}{Brenden~M Lake}, \bibinfo{person}{Tomer~D Ullman}, \bibinfo{person}{Joshua~B Tenenbaum}, {and} \bibinfo{person}{Samuel~J Gershman}.} \bibinfo{year}{2017}\natexlab{}.
\newblock \showarticletitle{Building machines that learn and think like people}.
\newblock \bibinfo{journal}{\emph{Behavioral and brain sciences}}  \bibinfo{volume}{40} (\bibinfo{year}{2017}), \bibinfo{pages}{e253}.
\newblock


\bibitem[Lee et~al\mbox{.}(2018)]%
        {Lee2018}
\bibfield{author}{\bibinfo{person}{Chung-Wei Lee}, \bibinfo{person}{Wei Fang}, \bibinfo{person}{Chih-Kuan Yeh}, {and} \bibinfo{person}{Yu-Chiang~Frank Wang}.} \bibinfo{year}{2018}\natexlab{}.
\newblock \showarticletitle{Multi-label Zero-Shot Learning with Structured Knowledge Graphs}. In \bibinfo{booktitle}{\emph{IEEE Conference on Computer Vision and Pattern Recognition}}. \bibinfo{pages}{1576--1585}.
\newblock


\bibitem[Lee et~al\mbox{.}(2007)]%
        {Lee2007}
\bibfield{author}{\bibinfo{person}{Honglak Lee}, \bibinfo{person}{Chaitanya Ekanadham}, {and} \bibinfo{person}{Andrew~Y. Ng}.} \bibinfo{year}{2007}\natexlab{}.
\newblock \showarticletitle{Sparse Deep Belief Net Model for Visual Area V2}. In \bibinfo{booktitle}{\emph{Neural Information Processing Systems}}. \bibinfo{pages}{873–880}.
\newblock


\bibitem[Li et~al\mbox{.}(2016)]%
        {Li2016}
\bibfield{author}{\bibinfo{person}{Qiang Li}, \bibinfo{person}{Maoying Qiao}, \bibinfo{person}{Wei Bian}, {and} \bibinfo{person}{Dacheng Tao}.} \bibinfo{year}{2016}\natexlab{}.
\newblock \showarticletitle{Conditional Graphical Lasso for Multi-label Image Classification}. In \bibinfo{booktitle}{\emph{IEEE Conference on Computer Vision and Pattern Recognition}}. \bibinfo{pages}{2977--2986}.
\newblock


\bibitem[Li et~al\mbox{.}(2014)]%
        {Li2014}
\bibfield{author}{\bibinfo{person}{Xin Li}, \bibinfo{person}{Feipeng Zhao}, {and} \bibinfo{person}{Yuhong Guo}.} \bibinfo{year}{2014}\natexlab{}.
\newblock \showarticletitle{Multi-label Image Classification with A Probabilistic Label Enhancement Model}. In \bibinfo{booktitle}{\emph{Conference on Uncertainty in Artificial Intelligence}} \emph{(\bibinfo{series}{UAI'14})}. \bibinfo{pages}{430–439}.
\newblock


\bibitem[Lin et~al\mbox{.}(2014)]%
        {Lin2014}
\bibfield{author}{\bibinfo{person}{Tsung-Yi Lin}, \bibinfo{person}{Michael Maire}, \bibinfo{person}{Serge Belongie}, \bibinfo{person}{James Hays}, \bibinfo{person}{Pietro Perona}, \bibinfo{person}{Deva Ramanan}, \bibinfo{person}{Piotr Doll{\'a}r}, {and} \bibinfo{person}{C.~Lawrence Zitnick}.} \bibinfo{year}{2014}\natexlab{}.
\newblock \showarticletitle{Microsoft COCO: Common Objects in Context}. In \bibinfo{booktitle}{\emph{European Conference on Computer Vision}}. \bibinfo{pages}{740--755}.
\newblock


\bibitem[Liu et~al\mbox{.}(2021)]%
        {Liu2021Q2L}
\bibfield{author}{\bibinfo{person}{Shilong Liu}, \bibinfo{person}{Lei Zhang}, \bibinfo{person}{Xiao Yang}, \bibinfo{person}{Hang Su}, {and} \bibinfo{person}{Jun Zhu}.} \bibinfo{year}{2021}\natexlab{}.
\newblock \showarticletitle{Query2Label: A Simple Transformer Way to Multi-Label Classification}.
\newblock \bibinfo{journal}{\emph{arXiv preprint arXiv:2107.10834}} (\bibinfo{year}{2021}).
\newblock


\bibitem[Long et~al\mbox{.}(2015)]%
        {Long2015}
\bibfield{author}{\bibinfo{person}{Jonathan Long}, \bibinfo{person}{Evan Shelhamer}, {and} \bibinfo{person}{Trevor Darrell}.} \bibinfo{year}{2015}\natexlab{}.
\newblock \showarticletitle{Fully Convolutional Networks for Semantic Segmentation}. In \bibinfo{booktitle}{\emph{IEEE Conference on Computer Vision and Pattern Recognition}}. \bibinfo{pages}{3431--3440}.
\newblock


\bibitem[Loshchilov and Hutter(2017)]%
        {loshchilov2017decoupled}
\bibfield{author}{\bibinfo{person}{Ilya Loshchilov} {and} \bibinfo{person}{Frank Hutter}.} \bibinfo{year}{2017}\natexlab{}.
\newblock \showarticletitle{Decoupled Weight Decay Regularization}.
\newblock \bibinfo{journal}{\emph{arXiv preprint arXiv:1711.05101}} (\bibinfo{year}{2017}).
\newblock


\bibitem[Mittal et~al\mbox{.}(2024)]%
        {mittal2024}
\bibfield{author}{\bibinfo{person}{Trisha Mittal}, \bibinfo{person}{Sanjoy Chowdhury}, \bibinfo{person}{Pooja Guhan}, \bibinfo{person}{Snikitha Chelluri}, {and} \bibinfo{person}{Dinesh Manocha}.} \bibinfo{year}{2024}\natexlab{}.
\newblock \showarticletitle{Towards Determining Perceived Audience Intent for Multimodal Social Media Posts Using the Theory of Reasoned Action}.
\newblock \bibinfo{journal}{\emph{Scientific Reports}} \bibinfo{volume}{14}, \bibinfo{number}{1} (\bibinfo{year}{2024}), \bibinfo{pages}{10606}.
\newblock


\bibitem[Ren et~al\mbox{.}(2017)]%
        {Ren2017}
\bibfield{author}{\bibinfo{person}{Shaoqing Ren}, \bibinfo{person}{Kaiming He}, \bibinfo{person}{Ross Girshick}, {and} \bibinfo{person}{Jian Sun}.} \bibinfo{year}{2017}\natexlab{}.
\newblock \showarticletitle{Faster R-CNN: Towards Real-Time Object Detection with Region Proposal Networks}.
\newblock \bibinfo{journal}{\emph{IEEE Transactions on Pattern Analysis and Machine Intelligence}} \bibinfo{volume}{39}, \bibinfo{number}{6} (\bibinfo{year}{2017}), \bibinfo{pages}{1137–1149}.
\newblock


\bibitem[Ridnik et~al\mbox{.}(2021)]%
        {ridnik2021asymmetric}
\bibfield{author}{\bibinfo{person}{Tal Ridnik}, \bibinfo{person}{Emanuel Ben-Baruch}, \bibinfo{person}{Nadav Zamir}, \bibinfo{person}{Asaf Noy}, \bibinfo{person}{Itamar Friedman}, \bibinfo{person}{Matan Protter}, {and} \bibinfo{person}{Lihi Zelnik-Manor}.} \bibinfo{year}{2021}\natexlab{}.
\newblock \showarticletitle{Asymmetric Loss for Multi-label Classification}. In \bibinfo{booktitle}{\emph{IEEE International Conference on Computer Vision}}. \bibinfo{pages}{82--91}.
\newblock


\bibitem[Selvaraju et~al\mbox{.}(2017)]%
        {selvaraju2017}
\bibfield{author}{\bibinfo{person}{Ramprasaath~R Selvaraju}, \bibinfo{person}{Michael Cogswell}, \bibinfo{person}{Abhishek Das}, \bibinfo{person}{Ramakrishna Vedantam}, \bibinfo{person}{Devi Parikh}, {and} \bibinfo{person}{Dhruv Batra}.} \bibinfo{year}{2017}\natexlab{}.
\newblock \showarticletitle{Grad-Cam: Visual Explanations from Deep Networks via Gradient-Based Localization}. In \bibinfo{booktitle}{\emph{IEEE International Conference on Computer Vision}}. \bibinfo{pages}{618--626}.
\newblock


\bibitem[Shi et~al\mbox{.}(2024)]%
        {Shi2024}
\bibfield{author}{\bibinfo{person}{QingHongYa Shi}, \bibinfo{person}{Mang Ye}, \bibinfo{person}{Wenke Huang}, \bibinfo{person}{Weijian Ruan}, {and} \bibinfo{person}{Bo Du}.} \bibinfo{year}{2024}\natexlab{}.
\newblock \showarticletitle{Label-Aware Calibration and Relation-Preserving in Visual Intention Understanding}.
\newblock \bibinfo{journal}{\emph{IEEE Transactions on Image Processing}}  \bibinfo{volume}{33} (\bibinfo{year}{2024}), \bibinfo{pages}{2627--2638}.
\newblock


\bibitem[Shi et~al\mbox{.}(2023)]%
        {Shi2023}
\bibfield{author}{\bibinfo{person}{QingHongYa Shi}, \bibinfo{person}{Mang Ye}, \bibinfo{person}{Ziyi Zhang}, {and} \bibinfo{person}{Bo Du}.} \bibinfo{year}{2023}\natexlab{}.
\newblock \showarticletitle{Learnable Hierarchical Label Embedding and Grouping for Visual Intention Understanding}.
\newblock \bibinfo{journal}{\emph{IEEE Transactions on Affective Computing}} \bibinfo{volume}{14}, \bibinfo{number}{4} (\bibinfo{year}{2023}), \bibinfo{pages}{3218–3230}.
\newblock


\bibitem[Snell et~al\mbox{.}(2017)]%
        {snell2017prototypical}
\bibfield{author}{\bibinfo{person}{Jake Snell}, \bibinfo{person}{Kevin Swersky}, {and} \bibinfo{person}{Richard Zemel}.} \bibinfo{year}{2017}\natexlab{}.
\newblock \showarticletitle{Prototypical Networks for Few-shot Learning}. In \bibinfo{booktitle}{\emph{Neural information processing systems}}, Vol.~\bibinfo{volume}{30}.
\newblock


\bibitem[Wang et~al\mbox{.}(2023)]%
        {Wang2023}
\bibfield{author}{\bibinfo{person}{Binglu Wang}, \bibinfo{person}{Kang Yang}, \bibinfo{person}{Yongqiang Zhao}, \bibinfo{person}{Teng Long}, {and} \bibinfo{person}{Xuelong Li}.} \bibinfo{year}{2023}\natexlab{}.
\newblock \showarticletitle{Prototype-Based Intent Perception}.
\newblock \bibinfo{journal}{\emph{IEEE Transactions on Multimedia}}  \bibinfo{volume}{25} (\bibinfo{year}{2023}), \bibinfo{pages}{8308–8319}.
\newblock


\bibitem[Wang et~al\mbox{.}(2016)]%
        {Wang2016}
\bibfield{author}{\bibinfo{person}{Jiang Wang}, \bibinfo{person}{Yi Yang}, \bibinfo{person}{Junhua Mao}, \bibinfo{person}{Zhiheng Huang}, \bibinfo{person}{Chang Huang}, {and} \bibinfo{person}{Wei Xu}.} \bibinfo{year}{2016}\natexlab{}.
\newblock \showarticletitle{CNN-RNN: A Unified Framework for Multi-label Image Classification}. In \bibinfo{booktitle}{\emph{IEEE Conference on Computer Vision and Pattern Recognition}}. \bibinfo{pages}{2285--2294}.
\newblock


\bibitem[Wen et~al\mbox{.}(2023)]%
        {wen2023}
\bibfield{author}{\bibinfo{person}{Changsong Wen}, \bibinfo{person}{Guoli Jia}, {and} \bibinfo{person}{Jufeng Yang}.} \bibinfo{year}{2023}\natexlab{}.
\newblock \showarticletitle{Dip: Dual Incongruity Perceiving Network for Sarcasm Detection}. In \bibinfo{booktitle}{\emph{IEEE Conference on Computer Vision and Pattern Recognition}}. \bibinfo{pages}{2540--2550}.
\newblock


\bibitem[Yang et~al\mbox{.}(2018)]%
        {yang2018robust}
\bibfield{author}{\bibinfo{person}{Hong-Ming Yang}, \bibinfo{person}{Xu-Yao Zhang}, \bibinfo{person}{Fei Yin}, {and} \bibinfo{person}{Cheng-Lin Liu}.} \bibinfo{year}{2018}\natexlab{}.
\newblock \showarticletitle{Robust Classification with Convolutional Prototype Learning}. In \bibinfo{booktitle}{\emph{IEEE Conference on Computer Vision and Pattern Recognition}}. \bibinfo{pages}{3474--3482}.
\newblock


\bibitem[Ye et~al\mbox{.}(2023)]%
        {Ye2023}
\bibfield{author}{\bibinfo{person}{Mang Ye}, \bibinfo{person}{Qinghongya Shi}, \bibinfo{person}{Kehua Su}, {and} \bibinfo{person}{Bo Du}.} \bibinfo{year}{2023}\natexlab{}.
\newblock \showarticletitle{Cross-Modality Pyramid Alignment for Visual Intention Understanding}.
\newblock \bibinfo{journal}{\emph{IEEE Transactions on Image Processing}}  \bibinfo{volume}{32} (\bibinfo{year}{2023}), \bibinfo{pages}{2190–2201}.
\newblock


\bibitem[Zhang et~al\mbox{.}(2022)]%
        {zhang2022use}
\bibfield{author}{\bibinfo{person}{Shu Zhang}, \bibinfo{person}{Ran Xu}, \bibinfo{person}{Caiming Xiong}, {and} \bibinfo{person}{Chetan Ramaiah}.} \bibinfo{year}{2022}\natexlab{}.
\newblock \showarticletitle{Use All the Labels: A Hierarchical Multi-label Contrastive Learning Framework}. In \bibinfo{booktitle}{\emph{IEEE Conference on Computer Vision and Pattern Recognition}}. \bibinfo{pages}{16660--16669}.
\newblock


\bibitem[Zhao et~al\mbox{.}(2021)]%
        {Zhao2021}
\bibfield{author}{\bibinfo{person}{Jiawei Zhao}, \bibinfo{person}{Ke Yan}, \bibinfo{person}{Yifan Zhao}, \bibinfo{person}{Xiaowei Guo}, \bibinfo{person}{Feiyue Huang}, {and} \bibinfo{person}{Jia Li}.} \bibinfo{year}{2021}\natexlab{}.
\newblock \showarticletitle{Transformer-based Dual Relation Graph for Multi-label Image Recognition}. In \bibinfo{booktitle}{\emph{IEEE International Conference on Computer Vision}}. \bibinfo{pages}{163--172}.
\newblock


\bibitem[Zhu et~al\mbox{.}(2017)]%
        {Zhu2017}
\bibfield{author}{\bibinfo{person}{Feng Zhu}, \bibinfo{person}{Hongsheng Li}, \bibinfo{person}{Wanli Ouyang}, \bibinfo{person}{Nenghai Yu}, {and} \bibinfo{person}{Xiaogang Wang}.} \bibinfo{year}{2017}\natexlab{}.
\newblock \showarticletitle{Learning Spatial Regularization with Image-Level Supervisions for Multi-label Image Classification}. In \bibinfo{booktitle}{\emph{IEEE Conference on Computer Vision and Pattern Recognition}}. \bibinfo{pages}{2027--2036}.
\newblock


\bibitem[Zhu et~al\mbox{.}(2021)]%
        {zhu2021prototype}
\bibfield{author}{\bibinfo{person}{Fei Zhu}, \bibinfo{person}{Xu-Yao Zhang}, \bibinfo{person}{Chuang Wang}, \bibinfo{person}{Fei Yin}, {and} \bibinfo{person}{Cheng-Lin Liu}.} \bibinfo{year}{2021}\natexlab{}.
\newblock \showarticletitle{Prototype Augmentation and Self-supervision for Incremental Learning}. In \bibinfo{booktitle}{\emph{IEEE Conference on Computer Vision and Pattern Recognition}}. \bibinfo{pages}{5871--5880}.
\newblock


\end{thebibliography}




\end{document}